\pdfoutput=1

\documentclass[11pt]{article}

\usepackage{EMNLP2022}

\usepackage{times}
\usepackage{latexsym}

\usepackage[T1]{fontenc}

\usepackage[utf8]{inputenc}

\usepackage{microtype}

\usepackage{inconsolata}

\usepackage{subcaption}

\usepackage{colortbl}
\definecolor{Gray}{gray}{0.85}
\definecolor{aliceblue}{rgb}{0.94, 0.97, 1.0}
\definecolor{beaublue}{rgb}{0.74, 0.83, 0.9}
\definecolor{blond}{rgb}{0.98, 0.94, 0.75}
\definecolor{beige}{rgb}{0.96, 0.96, 0.86}
\definecolor{cornsilk}{rgb}{1.0, 0.97, 0.86}
\definecolor{platinum}{rgb}{0.9, 0.89, 0.89}
\definecolor{blue(pigment)}{rgb}{0.2, 0.2, 0.6}
\definecolor{goldenbrown}{rgb}{0.6, 0.4, 0.08}

\usepackage{enumerate}[leftmargin=*]
\usepackage{enumitem}
\setlist[itemize]{leftmargin=*}
\usepackage{tabularx}
\usepackage{booktabs,ragged2e}
\newcolumntype{Y}{>{\RaggedRight\arraybackslash}X}
\newcolumntype{L}[1]{>{\raggedright\arraybackslash}p{#1}}
\newcolumntype{C}[1]{>{\centering\arraybackslash}p{#1}}
\newcolumntype{M}[1]{>{\centering\arraybackslash}m{#1}}
\newcolumntype{R}[1]{>{\raggedleft\arraybackslash}p{#1}}
\usepackage{graphicx,array}
\usepackage{multirow}
\usepackage{hhline}
\usepackage[ruled,vlined,linesnumbered,longend]{algorithm2e}
\usepackage{amsmath,bbm,amsfonts}
\usepackage{amsthm}

\usepackage{mathtools}
\theoremstyle{definition}

\usepackage{xspace}
\newcommand{\method}{\texttt{MedCLIP}\xspace}
\newif\ifnotopic

\newcommand{\bv}{\mathbf{v}}
\newcommand{\bx}{\mathbf{x}}
\newcommand{\bt}{\mathbf{t}}
\newcommand{\bl}{\mathbf{l}}

\title{\method: Contrastive Learning from Unpaired Medical Images and Text}

\author{Zifeng Wang\textsuperscript{\rm 1}, 
Zhenbang Wu\textsuperscript{\rm 1}, Dinesh Agarwal\textsuperscript{\rm 1,3}, Jimeng Sun\textsuperscript{\rm 1,2} \\
\textsuperscript{\rm 1}Department of Computer Science, University of Illinois Urbana-Champaign \\
\textsuperscript{\rm 2}Carle Illinois College of Medicine, University of Illinois Urbana-Champaign \\
\textsuperscript{\rm 3}Adobe \\
\texttt{\{zifengw2, zw12, jimeng\}@illinois.edu, diagarwa@adobe.com}}

\begin{document}
\maketitle

\begin{abstract}
Existing vision-text contrastive learning like CLIP \cite{radford2021learning} aims to match the paired image and caption embeddings while pushing others apart, which improves representation transferability and supports zero-shot prediction. However, medical image-text datasets are orders of magnitude below the general images and captions from the internet. Moreover, previous methods encounter many false negatives, i.e., images and reports from separate patients probably carry the same semantics but are wrongly treated as negatives. In this paper, we decouple images and texts for multimodal contrastive learning thus scaling the usable training data in a combinatorial magnitude with low cost. We also propose to replace the InfoNCE loss with semantic matching loss based on medical knowledge to eliminate false negatives in contrastive learning. We prove that \method is a simple yet effective framework: it outperforms state-of-the-art methods on zero-shot prediction, supervised classification, and image-text retrieval. Surprisingly, we observe that with only 20K pre-training data, \method wins over the state-of-the-art method (using $\approx200$K data) \footnote{Our code is available at \url{https://github.com/RyanWangZf/MedCLIP}.}. 
\end{abstract}
\section{Introduction}

Medical images such as X-rays, CTs, and MRIs are commonly used to diagnose, monitor, or treat medical conditions in clinical practice~\cite{medical_imaging}. 
With the rapid growth of medical images and the corresponding reports data, researchers have developed various deep learning models to support clinical decision making \cite{ccalli2021deep}.

Recently, large-scale image-text pre-training, e.g., CLIP~\cite{radford2021learning}, has achieved considerable successes in computer vision and natural language processing domains. CLIP is trained to predict the correct matching of a batch of images and text training examples. The joint-training of image and text representations on large-scale image-text pairs generates transferable representations and supports flexible downstream tasks. Inspired by success of CLIP, we believe the knowledge jointly learned from medical images and reports should be helpful for downstream clinical tasks.

However, adopting vision-text pre-training on medical domain is a non-trivial task due to (1) CLIP's~\cite{radford2021learning} data-hungry nature: CLIP is trained on a dataset of 400M image-text pairs collected from the internet, while the total number of publicly available medical images and reports is orders of magnitude below; and (2) specificity of medical images and reports: compared to general domains (e.g., "cats" v.s. "dog"), the differences within medical domains are more subtle and fine-grained (e.g., "pneumonia" v.s. "consolidation"). In a nutshell, it is necessary to (1) address the data insufficiency issue; and (2) capture the subtle yet crucial medical meanings.

\begin{figure}[t]
 \centering
    \includegraphics[width=\linewidth]{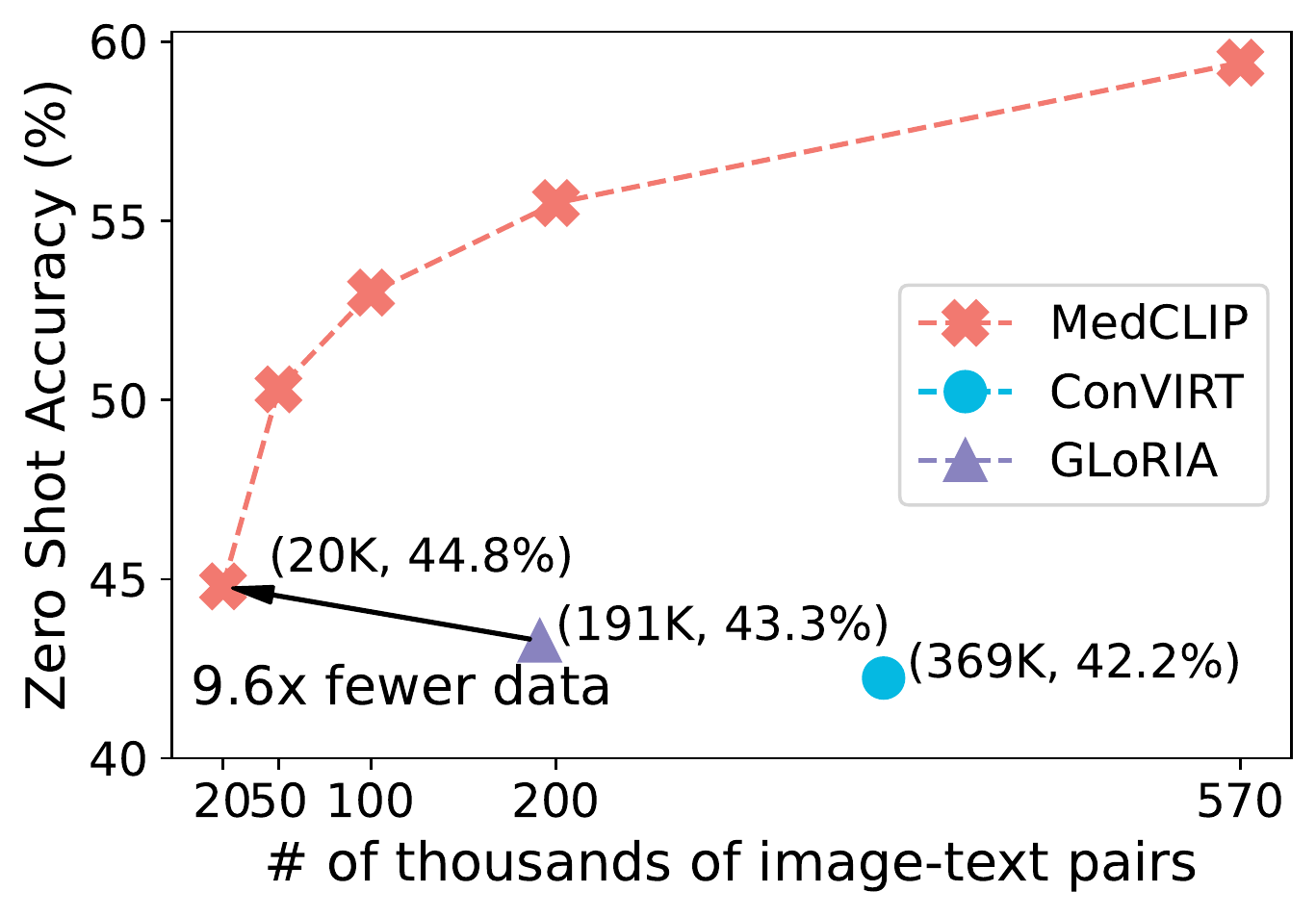}
    \caption{Zero-shot performance of \method, ConVIRT~\cite{zhang2020contrastive}, GLoRIA~\cite{huang2021gloria} when using different amounts of data for pre-training. ConVIRT and GLoRIA are trained on MIMIC-CXR (369K) and CheXpert (191K) dataset, respectively. Our method yields superior ACC than GLoRIA using near $1/10$ of pre-training data. }
    \label{fig:pre-training Data efficiency}
\end{figure}

\begin{figure*}[t]
    \centering
    \includegraphics[width=0.7\linewidth]{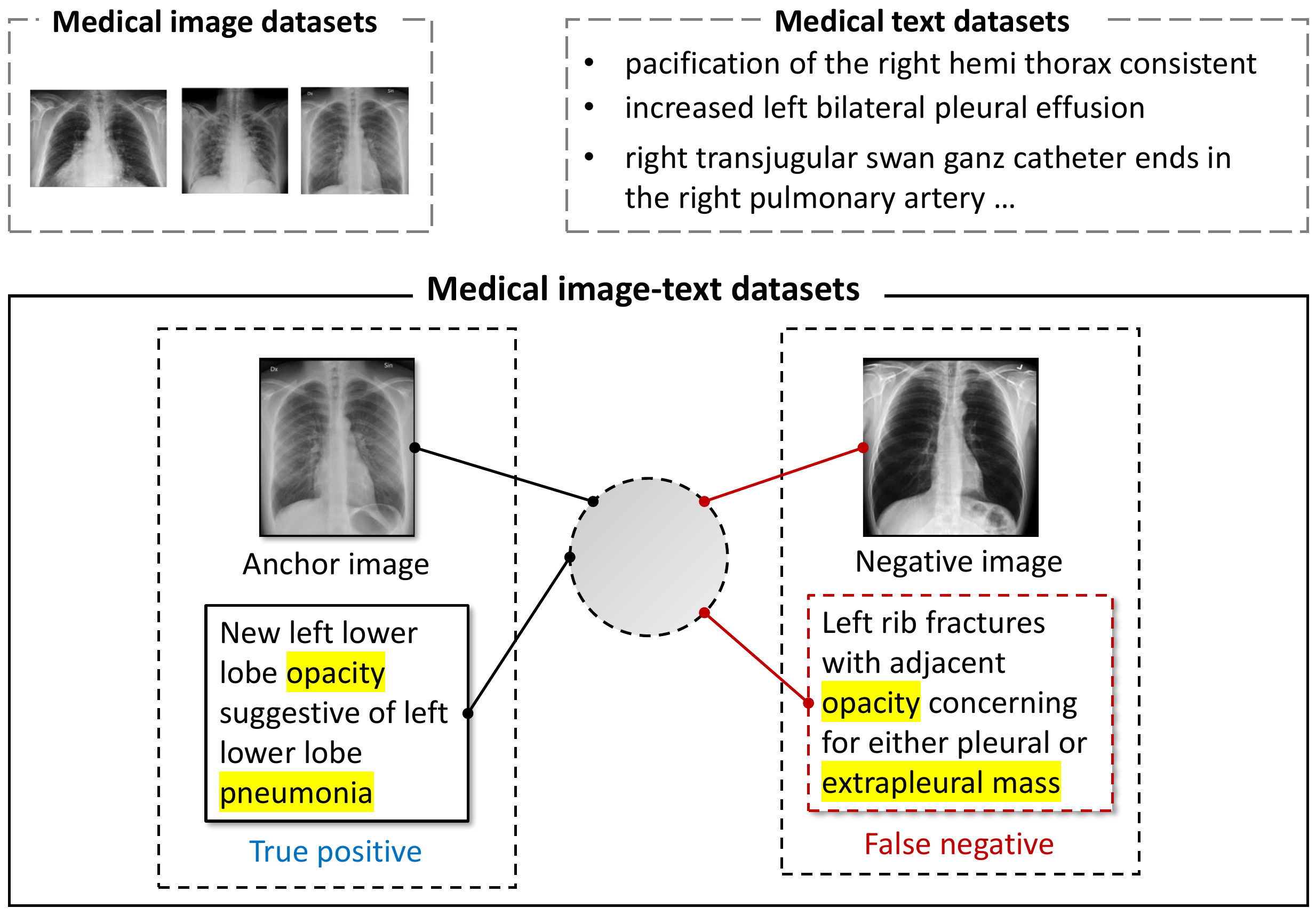}
    \caption{Demonstration of challenges in medical image-text contrastive learning. (1) Pre-training data only includes paired images and texts. However, many more image-only and text-only datasets are ignored. (2) False negatives appear. For an anchor image, previous methods treat paired texts (i.e., reports from the same patient's study) as positives and unpaired texts (i.e., reports from other patients' studies) as negatives. However, the negative texts can describe the same symptoms as the anchor texts.}
    \label{fig:demo}
\end{figure*}

Existing works try to tackle the challenges above in different ways. ConVIRT~\cite{zhang2020contrastive} jointly trains the vision and text encoders with the paired medical images and reports via a bidirectional contrastive objective; GLoRIA~\cite{huang2021gloria} further models both the global and local interactions between medical images and reports to capture the pathology meanings from specific image regions. However, both works have significant limitations, as illustrated in Fig. \ref{fig:demo}.
\begin{itemize}
    \item {\bf Limited usable data.} Most medical image datasets only provide the diagnostic labels instead of the raw reports. However, both works need paired image and reports, leaving a vast number of medical image-only and text-only datasets unused.
    \item {\bf False negatives in contrastive learning.} Both methods try to push images and texts embeddings from different patients apart. However, even though some reports do not belong to the target patient's study, they can still describe the same symptoms and findings. Simply treating the other reports as negative samples brings noise to the supervision and confuses the model.
\end{itemize}

To handle the above challenges, we propose a simple yet effective approach, namely \method. It has the following contributions:
\begin{itemize}
    \item {\bf Decoupling images and texts for contrastive learning.} We extend the pre-training to cover the massive unpaired images and texts datasets, which scales the number of training data in a combinatorial manner. It opens a new direction to expand multi-modal learning based on medical knowledge instead of expensively scaling up data.
    \item {\bf Eliminating false negatives via medical knowledge.} We observe that images and reports from separate patients' studies may carry the same semantics but are falsely treated as negatives by previous methods. Hence, we design a soft semantic matching loss that uses the medical semantic similarity between each image and report as the supervision signal. This approach equips the model with the ability to capture the subtle yet crucial medical meanings.
\end{itemize}

We make comprehensive evaluation on \method across four public datasets. Results show that \method reaches extremely high data efficiency, as shown in Fig. \ref{fig:pre-training Data efficiency}. Our method obtains better performances than the state-of-the-art GLoRIA~\cite{huang2021gloria} using only 10\% pre-training data. Extensive experiments verify \method's transferability to various downstream tasks. It wins over baselines by a large margin: over 10\% improvement of prediction ACC for zero-shot prediction and supervised image classification tasks on average; over 2\% improvement of retrieval precision. Details are in \S \ref{sec:experiment}. 

\section{Related Works}
Vision-text representation learning was shown to learn good visual representations \cite{joulin2016learning,li2017learning,sariyildiz2020learning,desai2021virtex,kim2021vilt,wang2021vlmo}. But all of them work on paired image and captions from general domain, e.g., Flickr \cite{joulin2016learning} and COCO Captions \cite{desai2021virtex}. Likewise, these methods do not support cross-modal retrieval hence do not support zero-shot predictions either.

Many propose to learn visual-semantic embedding for vision-text retrieval \cite{liu2019aligning,wu2019unified,lu2019vilbert,huang2020pixel,chen2021learning} by attention or objection detection models; and by vision-text contrastive learning \cite{zhang2020contrastive,jia2021scaling,yuan2021florence,yu2022coca} or multiple vision and text supervision \cite{singh2021flava,li2022blip}. They all work on general domain where near infinite web images and captions are available, which dwarfs the scale of medical image-text data. This challenge hurdles the execution of self-supervised CL for large vision-text transformers. Though remedies like data augmentation \cite{li2021supervision} and knowledge graph \cite{shen2022k} were proposed, the magnitude of used data is still far larger than medical data.

Medical image-text representation learning was investigated based on contrastive learning as well
\cite{zhang2020contrastive,huang2021gloria,wang2021self}. Nonetheless, they all work on paired medical images and texts so still encounter the lacking data challenge. Moreover, they all suffer from the false negative noises when adopting noise contrastive estimation (NCE) \cite{van2018representation} to perform instance discrimination \cite{wu2018unsupervised}, which undermines the representation quality \cite{arora2019theoretical,zheng2021weakly}. Our work bridges the gap by making the full use of all available medical data to support medical image-text pre-training. And we harness medical knowledge tailored to eliminate false negatives in contrastive learning to improve the pre-training data efficiency.
\begin{figure*}[t]
    \centering
    \includegraphics[width=0.9\linewidth]{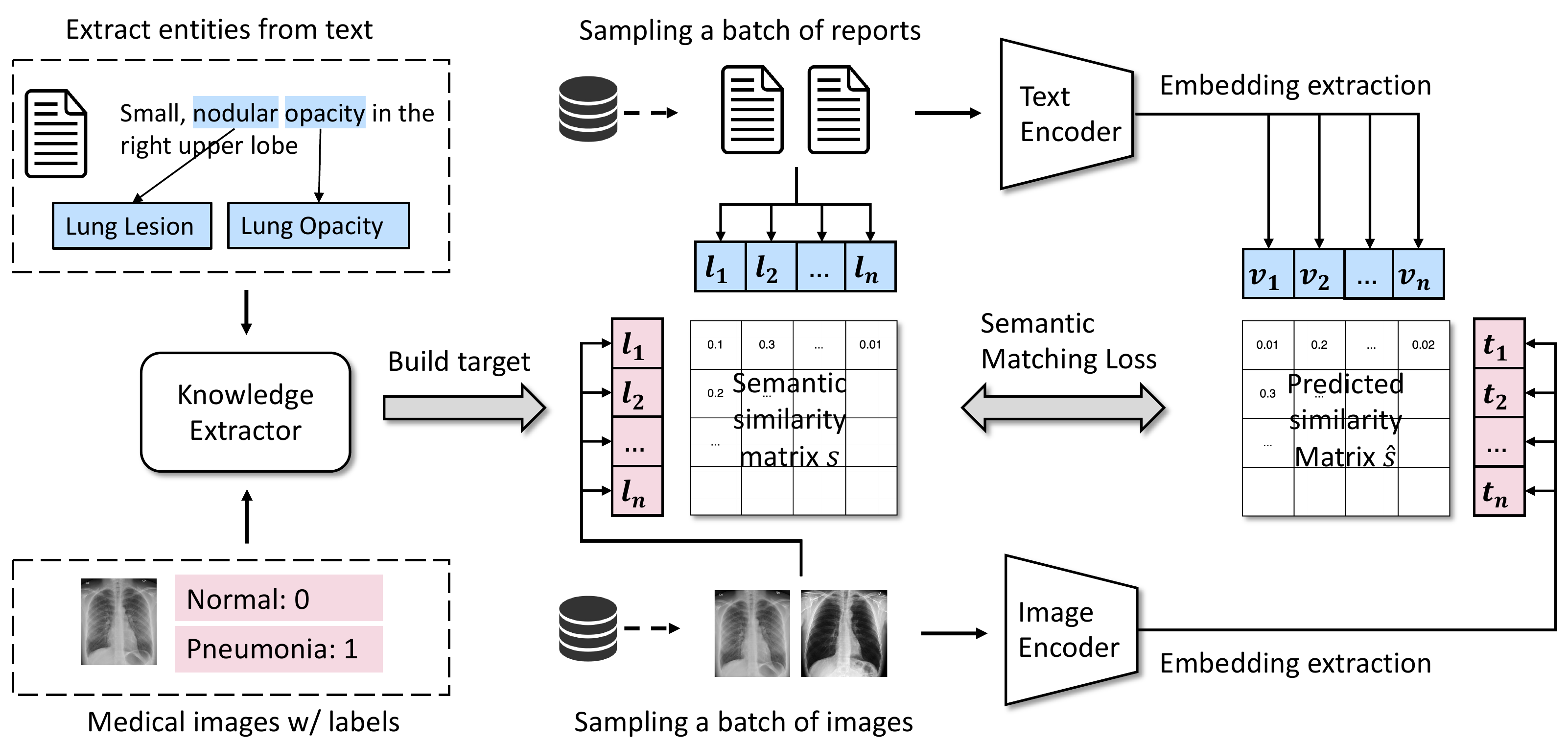}
    \caption{The workflow of \method. The knowledge extraction module extracts medical entities from raw medical reports. Then, a semantic similarity matrix is built by comparing medical entities (from text) and raw labels (from images), which enables pairing arbitrary two separately sampled images and texts. The extracted image and text embeddings are paired to match the semantic similarity matrix.   }
    \label{fig:architecture}
\end{figure*}

\section{Method}
In this section, we present the technical details of \method following the flow in Fig. \ref{fig:architecture}. \method consists of components (1) knowledge extraction that builds the \textit{semantic similarity matrix}, (2) vision and text encoders that extracts embeddings, and (3) \textit{semantic matching loss} that trains the whole model.

\subsection{Vision and Text Encoder}\label{sec:3.1}
\method consists of one visual encoder and one text encoder.\\

\noindent \textbf{Vision Encoder.} We encode images into embeddings $\bv \in\mathbb{R}^D$ using a vision encoder $E_{img}$. A projection head then maps raw embeddings to $\bv_p \in \mathbb{R}^P$.
\begin{subequations}
\begin{align}
\bv = E_{img}(\bx_{img}) \\
\bv_p = f_v(\bv)
\end{align} 
\end{subequations}
where $f_v$ is the projection head of the vision encoder.\\

\noindent \textbf{Text Encoder.} We create clinically meaningful text embeddings $\bt\in\mathbb{R}^M$ by a text encoder. We project them to $\bt_p\in\mathbb{R}^P$ as
\begin{subequations}
\begin{align}
\bt = E_{txt}(\bx_{txt}) \\
\bt_p = f_t(\bt)
\end{align} 
\end{subequations}
where $f_t$ is the projection head and $E_{txt}$ denotes the text encoder. This gives the same embedding dimension $P$ as the vision encoder, suitable for contrastive learning.

\subsection{Decouple Image-Text Pairs with Medical Knowledge Extractor}\label{sec:3.2}
Paired medical image text datasets are orders of magnitude less than the general paired image text (e.g., from the internet) due to the significant expense of supplying high-quality annotations by medical specialists as well as privacy and legal concerns. To enhance medical multi-modal learning, we want to make the full use of all existing medical image-text, image-only, and text-only datasets. The challenge is that for image-only, and text-only datasets, CLIP-like contrastive learning is infeasible. Also, we want to dig out all positive pairs to eliminate false negatives.

Suppose we have $n$ paired image-text samples, $m$ labeled images, and $h$ medical sentences. Previous methods are only able to use $n$ paired samples. By contrast, we decouple the $n$ paired samples into $n$ images and $n$ sentences, respectively. Ultimately, we are able to obtain $(n+m)\times (n+h)$ image-text pairs by traversing all possible combinations, which results in $\frac{(n+m)\times(n+h)}{n}\times$ more supervision. For instance, in Fig. \ref{fig:demo}, previous method pretrains on 2 image-text pairs while \method is capable of exploiting $(2+3)\times (2+3)=25$ samples in total.

To fulfill the additional supervision, we propose to leverage external medical knowledge to build the knowledge-driven \textit{semantic similarity}.  Unlike previous works that treat all positive samples equally \cite{khosla2020supervised,zheng2021weakly,wang2022transtab}, here we propose to differentiate samples via their semantic similarities.

In particular, we split raw reports into sentences $x_{txt}$. MetaMap \cite{aronson2010overview} is used to extract entities defined in Unified Medical Language System \cite{bodenreider2004unified} from raw sentences. Follow the practice of \citet{peng2018negbio}, we focus on 14 main entity types shown by Table ~\ref{tab:main_finding_type}. Likewise, for images with diagnosis labels, we leverage MetaMap to map the raw classes to UMLS conceptions thus being aligned with entities from texts, e.g., ``Normal'' maps to ``No Findings''. We build multi-hot vectors from the extracted entities for images and texts, as $\bl_{img}$ and $\bl_{txt}$, respectively. Therefore, we unify the semantics of images and texts. For any sampled $x_{img}$ and $x_{txt}$, we can measure their semantic similarity by comparing the corresponding $\bl_{img}$ and $\bl_{txt}$.

\subsection{Semantic Matching Loss} \label{sec:3.3}
We bridge the images and texts through the built semantic labels $\bl_{img}$ and $\bl_{txt}$. During each iteration, we sample $N_{batch}$ input images $\{\bx_{image}\}$ and text $\{\bx_{text}\}$ separately. Instead of defining positive pairing by searching equivalent labels, we propose to build soft targets $s$ by
\begin{equation}
    s = \frac{\bl_{img}^{\top} \cdot \bl_{txt}}{\|\bl_{img}\| \cdot \| \bl_{txt}\|}.
\end{equation}
$s$ thus indicates the medical semantic similarity.

For an image $i$, we obtain a set of $s_{ij}$ where $j=1\dots N_{batch}$ corresponds to the batch of texts. The soft target is computed by normalizing across $j$ by softmax.
\begin{equation}
y_{ij}^{v \rightarrow t} = \frac{\exp s_{ij}}{\sum_{j=1}^{N_{batch}} \exp s_{ij}}.
\end{equation}
Similarily, the reversed text-to-image soft targets are obtained by
\begin{equation}
y_{ji}^{t \rightarrow v} = \frac{\exp s_{ji}}{\sum_{i=1}^{N_{batch}} \exp s_{ji}}.
\end{equation}
The logits are obtained by cosine similarities between image and text embeddings:
\begin{equation}
    \hat{s}_{ij} = \tilde{\bv}_i^{\top} \cdot \tilde{\bt}_j,
\end{equation}
where $\tilde{\bv}_i$ and $\tilde{\bt}_j$ are normalized $\bv_p$ and $\bt_p$, respectively. The predicted similarity is also obtained by softmax function
\begin{equation}
    \hat{y}_{ij} =  \frac{\exp \hat{s}_{ij} / \tau}{\sum_{i=1}^{N_{batch}} \exp \hat{s}_{ij} / \tau}.
\end{equation}
$\tau$ is the temperature initialized at $0.07$. The \textit{semantic matching loss} is hence the cross entropy between the logits and soft targets as
\begin{equation}
  \mathcal{L}^{v \to l} = - \frac1{N_{batch}}\sum_{i=1}^{N_{batch}} \sum_{j=1}^{N_{batch}} y_{ij} \log \hat{y}_{ij}.
\end{equation}
Likewise, we can compute $\mathcal{L}^{l \to v}$ and then reach to 
\begin{equation}
    \mathcal{L} = \frac{\mathcal{L}^{v \to l}+\mathcal{L}^{l \to v}}{2}
\end{equation}
as the final training objective.

\begin{table*}[t]
  \centering
  \caption{Results of zero-shot image classification tasks on four datasets. We take an additional prompt ensemble version of each method (with subscript $_{\text{ENS}}$). We take the mean and standard deviation (STD) of accuracy (ACC) in five runs considering the randomness of prompt generation process. Best scores across a dataset are in bold.}
    \begin{tabular}{l|llll}
    \toprule
    ACC(STD) & CheXpert-5x200 & MIMIC-5x200 & COVID & RSNA \\
    \midrule
    CLIP & 0.2016(0.01) & 0.1918(0.01) & 0.5069(0.03) & 0.4989(0.01)\\
    CLIP$_{\text{ENS}}$ & 0.2036(0.01) & 0.2254(0.01) & 0.5090(<0.01) & 0.5055(0.01)\\
    \midrule
    ConVIRT & 0.4188(0.01) & 0.4018(0.01) & 0.5184(0.01) & 0.4731(0.05) \\
    ConVIRT$_{\text{ENS}}$ &  0.4224(0.02) & 0.4010(0.02) & 0.6647(0.05) & 0.4647(0.08) \\
    GLoRIA & 0.4328(0.01) & 0.3306(0.01) & 0.7090(0.04) & 0.5808(0.08) \\
    GLoRIA$_{\text{ENS}}$  & 0.4210(0.03) & 0.3382(0.01) & 0.5702(0.06) & 0.4752(0.06) \\
    \midrule 
    \rowcolor{platinum}
    \method-ResNet & 0.5476(0.01) & 0.5022(0.02) & \textbf{0.8472(<0.01)} & 0.7418(<0.01) \\
    \rowcolor{platinum}
    \method-ResNet$_{\text{ENS}}$  & 0.5712(<0.01) & \textbf{0.5430(<0.01)} & 0.8369(<0.01) & 0.7584(<0.01) \\
    \rowcolor{platinum}
    \method-ViT & 0.5942(<0.01) & 0.5006(<0.01) & 0.8013(<0.01) & 0.7447(0.01) \\
    \rowcolor{platinum}
    \method-ViT$_{\text{ENS}}$  & \textbf{0.5942(<0.01)} & 0.5024(<0.01) & 0.7943(<0.01) & \textbf{0.7682(<0.01)} \\
    \bottomrule
    \end{tabular}%
  \label{tab:zeroshot}%
\end{table*}%

\begin{table}[t]
  \centering
  \caption{Results of medical image classification tasks after fine-tuning. Best scores are in bold.}
   \resizebox{\linewidth}{!}{%
    \begin{tabular}{c|cccc}
    \toprule
    \multirow{2}{*}{ACC} & CheXpert & MIMIC  & \multirow{2}{*}{COVID}  & \multirow{2}{*}{RSNA} \\
    & -5x200   & -5x200 & & \\
    \midrule
    Random   & 0.2500 & 0.2220 & 0.5056 & 0.6421 \\
    ImageNet & 0.3200 & 0.2830 & 0.6020 & 0.7560 \\
    CLIP     & 0.3020 & 0.2780 & 0.5866 & 0.7303 \\
    \midrule
    ConVIRT & 0.4770  & 0.4040  & 0.6983 & 0.7846 \\
    GLoRIA  & 0.5370  & 0.3590  & 0.7623 & 0.7981 \\
    \rowcolor{platinum}
    \method & \textbf{0.5960} & \textbf{0.5650} & \textbf{0.7890} & \textbf{0.8075} \\
    \bottomrule
    \end{tabular}%
    }
  \label{tab:finetune}%
\end{table}%


\section{Experiments}\label{sec:experiment}
We conduct extensive experiments on four X-ray datasets to answer the following questions:
\begin{itemize}
    \item \textbf{Q1.} Does the proposed pre-training method yield better zero-shot image recognition performances?
    \item \textbf{Q2.} Does the knowledge-driven supervision, i.e., semantic matching task, facilitate the contrastive image-text pre-training? 
    \item \textbf{Q3.} Does \method bring better performance and label efficiency for downstream classification tasks with fine-tuning?
    \item \textbf{Q4.} Are the learned embeddings good at cross-modal retrieval tasks?
    \item \textbf{Q5.} How do the learned embeddings look like?
\end{itemize}

\begin{table}[t]
  \centering
  \caption{The statistics of used datasets. Pos.\%: positive sample ratio.}
   \resizebox{\linewidth}{!}{%
    \begin{tabular}{l|ccc}
    \toprule
    \textbf{Pretrain} & \textbf{\# Images} & \textbf{\# Reports} & \textbf{\# Classes} \\
    \midrule
    MIMIC-CXR & 377,111 & 201,063 & - \\
    CheXpert & 223,415 & -     & 14 \\
    \midrule
    \textbf{Evaluation} &\textbf{ \# Train (Pos.\%)} & \textbf{ \# Test (Pos.\%)} &\textbf{ \# Classes} \\
    \midrule
    CheXpert-5x200 & 1,000 (-)  &1,000 (-) & 5 \\
    MIMIC-5x200 & 1,000 (-)  &1,000 (-) & 5 \\
    COVID & 2,162 (19\%) & 3,000 (49\%) & 2\\
    RSNA & 8,486 (50\%) & 3,538 (50\%) & 2 \\
    \bottomrule
    \end{tabular}%
    }
  \label{tab:data_stats}%
\end{table}%

\subsection{Datasets}\label{appx:dataset}

\noindent\textbf{CheXpert}~\cite{irvin2019chexpert} is a large dataset of chest X-rays with 14 observation labels collected from Stanford Hospital. Note that this dataset does not provide the corresponding medical reports to the public. We use the training split of this dataset for pre-training. For evaluation, we follow \cite{huang2021gloria} and sample a multi-class classification dataset from the testing split, namely CheXpert-5x200. This multi-class classification dataset has 200 exclusively positive images for the five CheXpert competition tasks: Atelectasis, Cardiomegaly, Edema, Pleural, Effsion.\\

\noindent\textbf{MIMIC-CXR}~\cite{johnson2019mimic} is a large chest X-ray database with free-text radiology reports collected from the Beth Israel Deaconess Medical Center in Boston, MA. We use the training split of this dataset for pre-training. For evaluation, we also sample a MIMIC-5x200 dataset for the same five tasks above.\\

\noindent\textbf{COVID} \cite{rahman2021exploring} is a publicly available x-ray dataset with COVID v.s. non-COVID labels. The positive and negative ratio is roughly 1:1. We use this dataset for evaluation.\\

\noindent\textbf{RSNA Pneumonia} \cite{shih2019augmenting} is a collection of pneumonia cases found in the database of chest x-rays made public by the National Institutes of Health. This is a binary classification dataset: pneumonia v.s. normal. We sample a balanced subset (i.e., 1:1 positive and negative ratio) and use it for evaluation.\\

\subsection{Baselines}
\noindent\textbf{Random} is a ResNet-50~\cite{https://doi.org/10.48550/arxiv.1512.03385} model with its default random initialization.\\

\noindent\textbf{ImageNet} is a ResNet-50~\cite{https://doi.org/10.48550/arxiv.1512.03385} model with weights pretrained on the standard ImageNet ILSVRC-2012 task~\cite{5206848}.\\

\noindent\textbf{CLIP}~\cite{radford2021learning} is a vision-text contrastive learning framework pre-trained on a dataset of 400M image-texts pairs collected from the internet. \\

\noindent\textbf{ConVIRT} works on vision-text contrastive learning in medicine. It employs a plain InfoNCE loss \cite{van2018representation} on paired X-rays and reports. We reproduce it based on their paper based on BioClinicalBERT text encoder and ResNet50 \cite{he2016deep} vision encoder.\\

\noindent\textbf{GLoRIA} \cite{huang2021gloria} entangles image sub-regions and words in inference by cross-attention which was argued to better capture key characteristics in images and reports. We implement it based on the official code and the provided pretrained weights \footnote{\url{https://github.com/marshuang80/gloria}}.

\subsection{Implementation Details}\label{appx:implementation}
We use the BioClinicalBERT \footnote{\url{https://huggingface.co/emilyalsentzer/Bio_ClinicalBERT}} as the backbone text encoder and Swin Transformer~\cite{https://doi.org/10.48550/arxiv.2103.14030} with ImageNet~\cite{5206848} pre-trained weight as the backbone vision encoder. Both transformer-based models are drawn from the transformers library~\cite{wolf2019huggingface}. We also provide ablation study with ResNet-50~\cite{https://doi.org/10.48550/arxiv.1512.03385} as the vision encoder, which is in-line with previous works~\cite{zhang2020contrastive,huang2021gloria}.

MIMIC-CXR and CheXpert are used for pretraining where we held 5000 samples out out for evaluation. All images are padded to square then scaled to $224\times224$. For MIMIC-CXR, we combine the ``Findings'' and ``Impression'' sections of reports then split them into sentences. We remove sentences with less than 3 words.  We take a linear projection head with output dimension $512$, a learnable temperature $\tau$ initialized on $0.07$. We utilize image augmentations to first scale to raw images to $256\times256$ then apply random crop with size $224\times224$; horizontal flipping with $0.5$ probability; color jittering with brightness and contrast ratios from $[0.8,1.2]$; random affine transformation with degree sampled from $[-10,10]$, max translation rate $0.0625$, and scale factor in $[0.8,1.1]$. Other hyperparameters are: learning rate 5e-5, batch size $100$, weight decay 1e-4,  number of epochs $10$, learning rate warmup ratio $0.1$. We employ mixed-precision training such that the pretraining finishes in 8 hours on a single RTX-3090 GPU.

\subsection{Q1. Zero-Shot Classification}
We conduct zero-shot image classification evaluation on four datasets: CheXpert-5x200, MIMIC-5x200, COVID, and RSNA. The learned image-text encoders are used to support zero-shot prediction by matching the encoded image embeddings and the embeddings of created prompts for each disease class. We illustrate the results in Table \ref{tab:zeroshot}.

It can be found that our method outperforms all the other baselines by a great margin. \method is capable of benefiting from prompt ensemble to yield better performance. By contrast, the ensemble does not always lead to positive effect to the other two, especially that GLoRIA is usually harmed by ensemble. One reason might be that ConVIRT and GLoRIA cannot differentiate false negatives in contrastive pre-training, and those false negatives are incoporated with prompt ensemble which confuses the model.
Besides, we observe that the original CLIP model yields bad predictions that are basically identical to random guess on all datasets. It demonstrates the discrepancy between the general internet image-text and the ones in medical domain.

Interestingly, \method yields over 0.8 ACC on COVID data while there is no COVID-19 positive image available during the course of pre-training. To endow the model to detect COVID-19 infection, we refer to the descriptions proposed in \cite{smith2020characteristic}: \textit{the presence of patchy or confluent, bandlike ground-glass opacity or consolidation in a peripheral and mid to lower lung zone distribution}, to build the prompts. This result demonstrates that contrastive pre-training of \method provides it with the transferability to out-of-domain classes.

\subsection{Q2. Pre-training Data Efficiency}
Data efficiency is a key challenge in CLIP based methods. As evident, CLIP uses 400M image-text pairs in the training phase, which is not just computationally expensive but also infeasible in medical domain due to limited data. To evaluate the data efficiency of \method, we subsample the pre-training data to 20K, 50K, and 200K, then pre-train \method and record the yielded model zero-shot prediction on CheXpert-5x200 data. Results show in Fig. \ref{fig:pre-training Data efficiency}.

We surprisingly find that with 20K data \method yields superior performance over GLoRIA that learns from the whole CheXpert dataset (around 200K image-text pairs). Likewise, \method beats ConVIRT that uses 369K data. When we include more training data, \method obtains a lift on its accuracy as well. We do not observe the saturation of zero-shot ACC at 570K samples (MIMIC-CXR plus CheXpert). It signifies the great capacity of \method on learning from multi-sourced data.

\begin{figure}[t]
\begin{subfigure}[t]{\linewidth}
    \includegraphics[width=\linewidth]{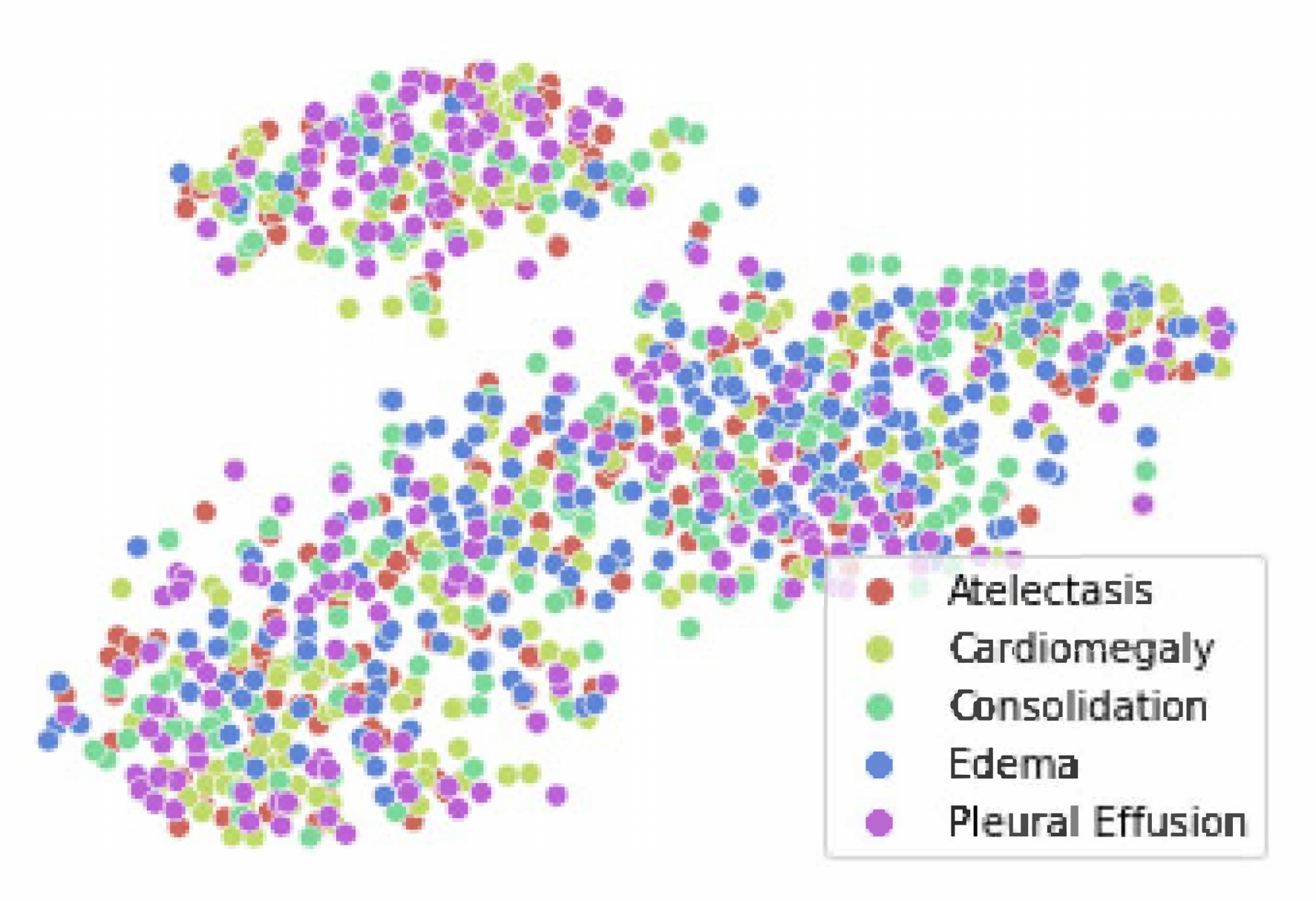}
    \caption{CLIP}
\end{subfigure}
\begin{subfigure}[t]{\linewidth}
    \includegraphics[width=\linewidth]{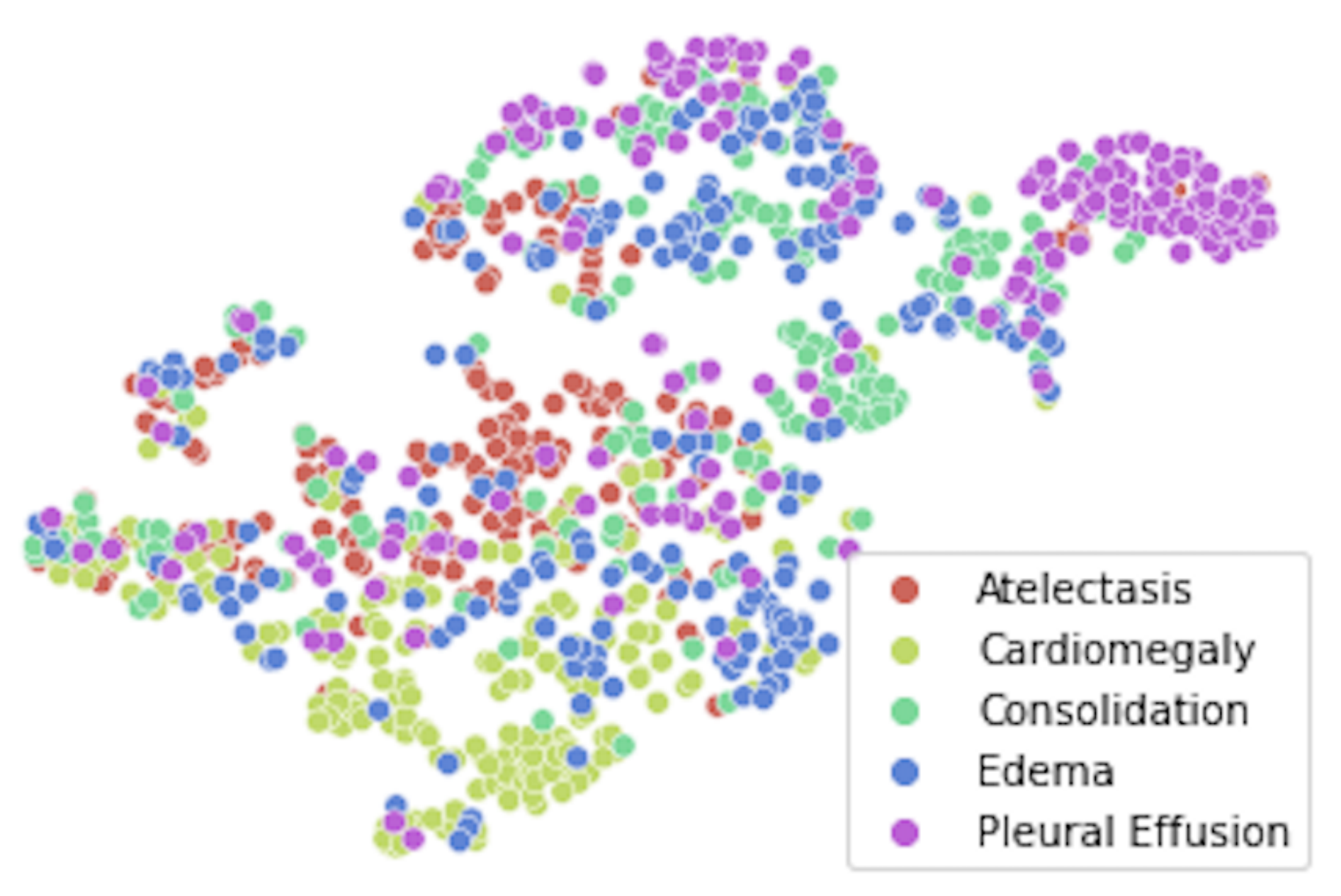}
    \caption{\method}
\end{subfigure}
\caption{Embeddings visualization of CheXpert5x200 images by CLIP and \method. Dimension reduced by t-SNE.}
    \label{fig:embedding-visualisation}
\end{figure}

\subsection{Q3. Fine-tune for Classification}
We aim to evaluate the learned model transferability to downstream supervised tasks. We draw and froze the image encoder and fine-tune a randomly initialized linear classification head on the training data with cross-entropy loss. Results are in Table \ref{tab:finetune}. We show that \method still achieves the best performances across all three methods. What is more, we surprisingly find that \method makes zero-shot prediction comparable with supervised learning models when 
contrasting Table \ref{tab:finetune} to Table \ref{tab:zeroshot}. 

On all datasets, the zero-shot \method performs better than the other supervised models. Specifically, we find on COVID, zero-shot \method performs better than its supervised counterpart, which demonstrates the supremacy of \method on low-resource scenarios. On the contrary, without pre-training on medical domain data, the ResNet baselines reach inferior performances. 

\subsection{Q4. Image-Text Retrieval}\label{subs:Image-text retrieval}
\begin{table}[t]
  \centering
  \caption{Results of Image-Text retrieval tasks on CheXpert5x200 dataset. We take the Precision@\{1,2,5,10\} to measure the performance of various models in this task. Best within the data are in bold.}
    \begin{tabular}{c|cccc}
    \toprule
    Model & P@1 & P@2 & P@5 & P@10\\
    \midrule
    CLIP & 0.21 & 0.20 & 0.20 & 0.19 \\
    ConVIRT & 0.20 & 0.20 & 0.20 & 0.21 \\
    GLoRIA & \textbf{0.47} & 0.47 & 0.46 & 0.46 \\
    \rowcolor{platinum}
    \method & 0.45 & \textbf{0.49} & \textbf{0.48} & \textbf{0.50} \\
    \bottomrule
    \end{tabular}%
  \label{tab:imag-text}%
\end{table}

We choose CheXpert-5x200 to evaluate the semantic richness of the learned representations by all models through the image-text retrieval task. Since CheXpert-5x200 do not have report data publicly available, we used MIMIC-CXR dataset to come up with reports/sentences. We sampled 200 sentences for each of the 5 classes as present in CheXpert-5x200 dataset. This gives rise to 1,000 images and 1,000 sentences as the retrieval dataset. We use Precision@\textit{K} to calculate the precision in the top \textit{K} retrieved reports/sentences by checking if the report belongs to the same category as the query image. 

We display the results by Table \ref{tab:imag-text}. It can be seen that \method achieves the best performances across all methods. This indicates that our method efficiently provide the required semantic information to retrieve texts. We find that there is an increase precision for \method with the higher \textit{K}. Analysis of this phenomenon is present in the Appendix \ref{appx:retrieval}.

\subsection{Q5. Embedding Visualization}
We also demonstrate the effectiveness of our representation learning framework by plotting t-SNE \cite{van2008visualizing} of image embeddings produced for CheXpert-5x200 images. We compare its embeddings with CLIP model embeddings. 
As visible in Fig. \ref{fig:embedding-visualisation}, our model produces better clustered representation. Whereas, CLIP model t-SNE plot is homogeneous. It is because most medical X-Rays share pretty high overlapping while only small lesion regions are different. Nonetheless, \method still detects clusters by the lesion types. 
\section{Conclusion}
In this work, we propose a decoupled medical image-text contrastive learning framework named \method. It significantly expands the training data size with a combinatorial magnitude. Meanwhile, the introduction of medical knowledge sheds light on alleviating false negatives. As a result, \method yields an excellent pretraining data efficiency: it wins over the state-of-the-art baseline by 1\% ACC with around $10\times$ fewer data. Moreover, we verify the prominence of \method on zero-shot prediction, supervised classification, and image-text retrieval tasks. It is expected to support a foundational model for the medical domain and handle medical diagnosis when facing diverse diseases with low resource requirements.

\section*{Acknowledgement}
This work was supported by NSF award SCH-2205289, SCH-2014438, IIS-1838042, NIH award R01 1R01NS107291-01.

\section*{Limitations}
This work leverages medical domain knowledge to decouple contrastive learning on medical images and texts. Hence, it significantly expands the available training data for pretraining. Meanwhile, the proposed knowledge-guided semantic matching loss debugs the false negatives appearing in naive contrastive learning.  It still encounters failure cases where incorrect semantic tags are detected or missing detection of negation or uncertainty phrases. A remedy can be introducing learning from noisy data techniques \cite{wang2020less,wang2021pico} to alleviate the noises in the extracted semantic similarity matrix.

Another concern is that though we prove \method is able to reach comparable zero-shot prediction accuracy to the finetuned counterpart, it is still not amenable to practical use. We suppose the reasons include (1) the prompt-based inference relies on the prompt quality and (2) more pretraining data is desired to further enhance the pretraining. Specifically, for the (1) point, it is promising to leverage prompt-learning methods \cite{zhou2022conditional} to automate the model application to downstream tasks instead of executing manual prompt engineering.

\bibliography{anthology,custom}
\bibliographystyle{acl_natbib}

\clearpage
\appendix
\begin{table}[t]
  \centering
  \caption{14 main finding types used in this paper.}
    \begin{tabular}{c}
    \toprule   
     Finding types \\
    \midrule   
     No Finding \\
     Enlarged Cardiomediastinum\\
     Cardiomegaly \\
     Lung Opacity \\
     Lung Lesion \\
     Edema  \\
     Consolidation \\
     Pneumonia  \\
     Atelectasis \\
     Pneumothorax \\
     Pleural Effusion \\
     Pleural Other \\
     Fracture \\
     Support Devices \\
    \bottomrule
    \end{tabular}%
  \label{tab:main_finding_type}%
\end{table}%

\begin{figure}[t]
\begin{subfigure}[t]{\linewidth}
    \includegraphics[width=\textwidth]{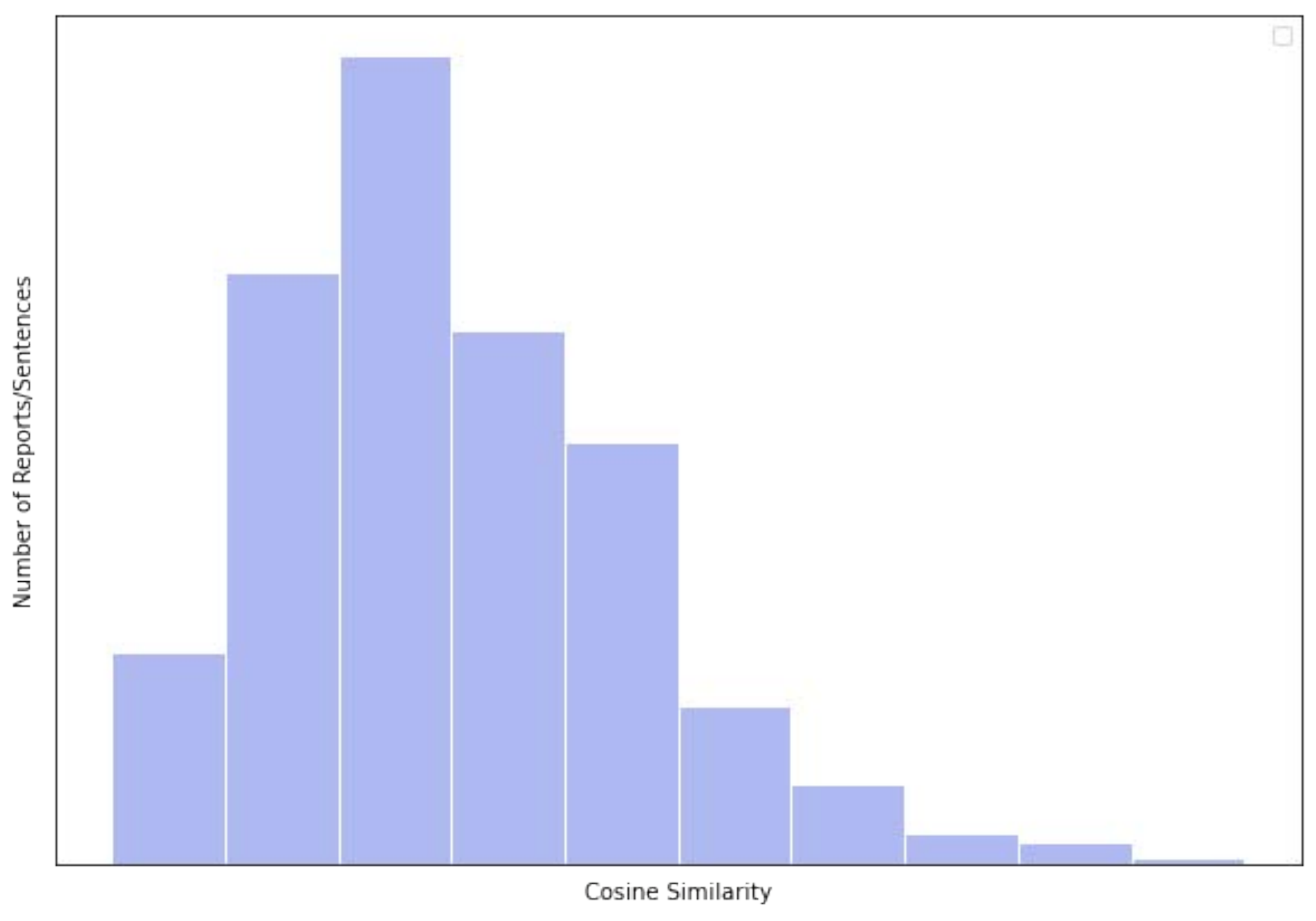}
    \caption{Histogram Plot for Atelectasis class}
\end{subfigure}
\begin{subfigure}[t]{\linewidth}
    \includegraphics[width=\textwidth]{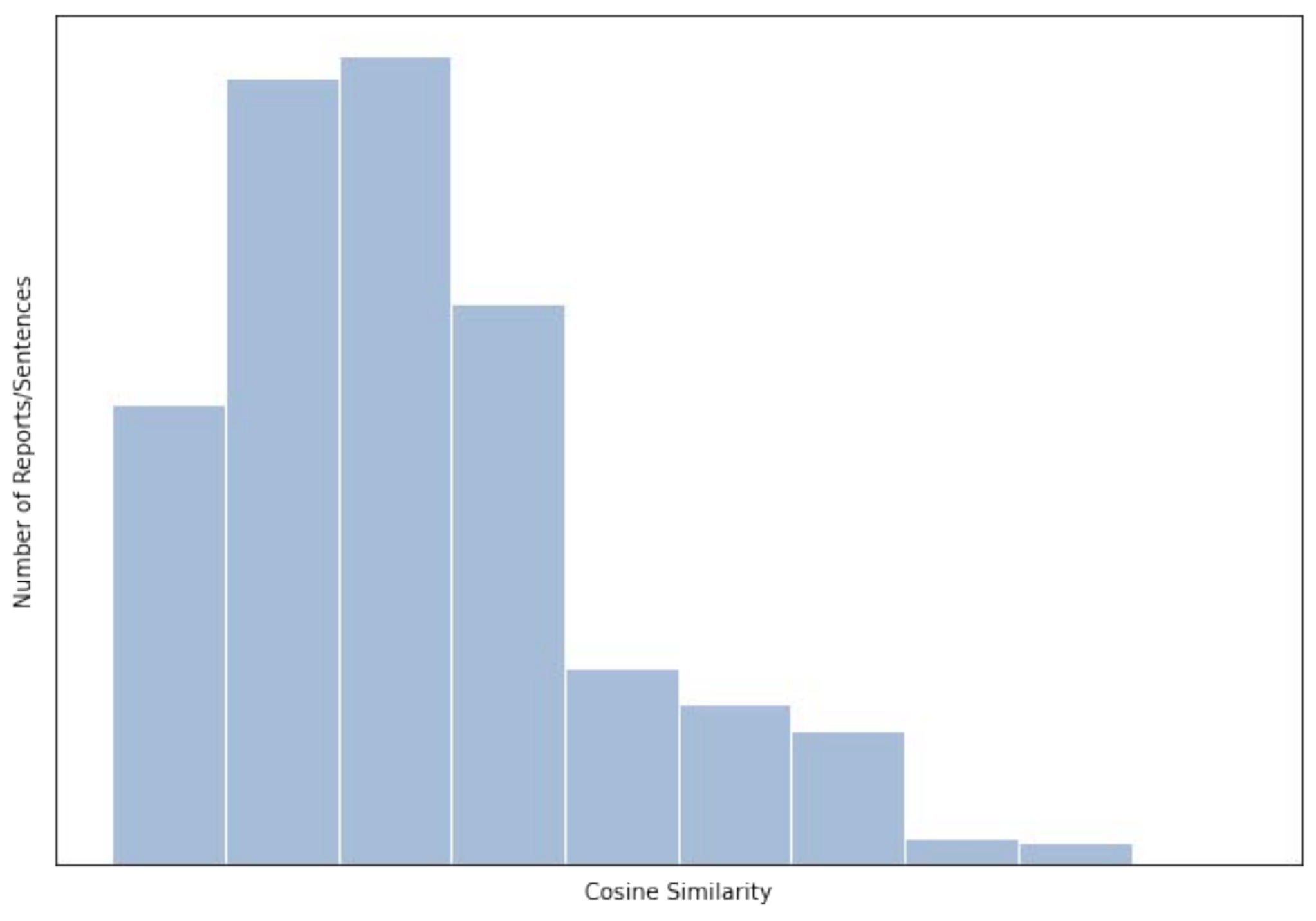}
    \caption{Histogram Plot for Cardiomegaly class}
\end{subfigure}
\begin{subfigure}[t]{\linewidth}
    \includegraphics[width=\textwidth]{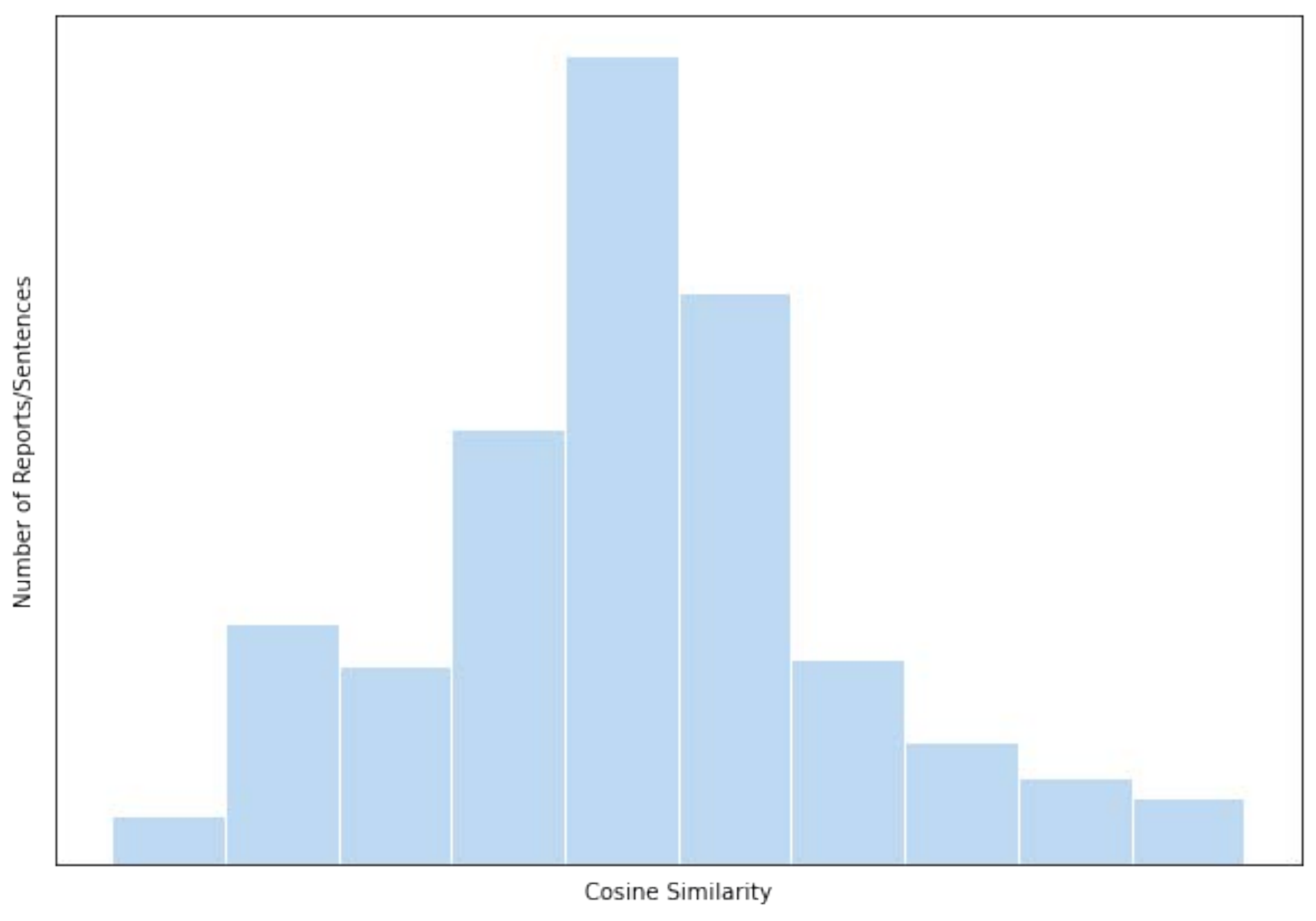}
    \caption{Histogram Plot for Consolidation class}
\end{subfigure}
\caption{Visualization of the similarity distributions computed based on \method embeddings.     \label{fig:image-text-retrieval-analysis}}
\end{figure}

\section{Analysis of Image-text retrieval results}\label{appx:retrieval}
We observed that there is certain increase in \method Precison@\textit{K} metric in Table \ref{tab:imag-text}. This phenomenon is sort of counterintuitive. As mentioned in Section \ref{subs:Image-text retrieval}, we used CheXpert-5x200 dataset for images and MIMIC-CXR for text in our image-text retrieval task. Each of them, i.e. images and text reports/sentences, has 1000 rows. Particularly, 200 images and sentences/reports are there for each class in CheXpert-5x200.

After running the main experiment we decided to plot some graphs to gain insight. Settings for the same is described as following: We pick up a class (e.g. Atelectasis) and for each image in that class we pick up sentences/reports, from top 10 (based on cosine distance) out of all the texts retrieved, that belong to the same class as that of the image in consideration. We also pick up their cosine similarity score. Finally, we plot a histogram (for each class) where x-axis is the cosine distance and the height of the bins of histogram represent the number of texts(retrieved in previous step) that have cosine similarity score of the bin. We have plotted few such plots and can be seen in Figure \ref{fig:image-text-retrieval-analysis}.


Intuitively, we are plotting number of texts that would appear in Precision@\textit{K} of a particular class given a cosine similarity threshold. Now as the histogram tells, there are more text centered around relatively smaller cosine similarity score. Further, as the \textit{K} increases in the Precision@\textit{K}, intuitively cut-off cosine similarity(or threshold) score would decrease and hence more text from the same class start showing up in the result. This explains why Precision@\textit{K} increased. In this sense, it is beneficial to investigate to address the non-smooth anisotropic distribution of image and text embeddings.

\end{document}